\documentclass[conference]{IEEEtran}
\IEEEoverridecommandlockouts
\usepackage{cite}
\usepackage{amsmath,amssymb,amsfonts}
\usepackage{algorithmic}
\usepackage{graphicx}
\usepackage{mathtools}
\usepackage{textcomp}
\usepackage{xcolor}
\usepackage{tabularx}
\usepackage{multirow}

\def\BibTeX{{\rm B\kern-.05em{\sc i\kern-.025em b}\kern-.08em
    T\kern-.1667em\lower.7ex\hbox{E}\kern-.125emX}}
\begin{document}

\title{Transformers for Green Semantic Communication: Less Energy, More Semantics}

\author{\IEEEauthorblockN{Shubhabrata Mukherjee, Cory Beard, and Sejun Song}
\IEEEauthorblockA{\textit{School of Science and Engineering, University of Missouri-Kansas City, Kansas City, MO, USA}
 \\
smpw5,beardc,songsej@umsystem.edu}
}

\maketitle

\begin{abstract}
Semantic communication aims to transmit meaningful and effective information, rather than focusing on individual symbols or bits. This results in benefits like reduced latency, bandwidth usage, and higher throughput compared with traditional communication. However, semantic communication poses significant challenges due to the need for universal metrics to benchmark the joint effects of semantic information loss and practical energy consumption. This research presents a novel multi-objective loss function named ``Energy-Optimized Semantic Loss'' (EOSL), addressing the challenge of balancing semantic information loss and energy consumption. Through comprehensive experiments on transformer models, including CPU and GPU energy usage, it is demonstrated that EOSL-based encoder model selection can save up to 90\% of energy while achieving a 44\% improvement in semantic similarity performance during inference in this experiment. This work paves the way for energy-efficient neural network selection and the development of greener semantic communication architectures.
\end{abstract}

\begin{IEEEkeywords}
Green semantic communication, Transformer, Energy optimized deep learning, Energy optimized loss function, Large language models

\end{IEEEkeywords}

\section{Introduction}

\begin{figure*}[hbt!]
\centerline{\includegraphics[width=0.95\textwidth]{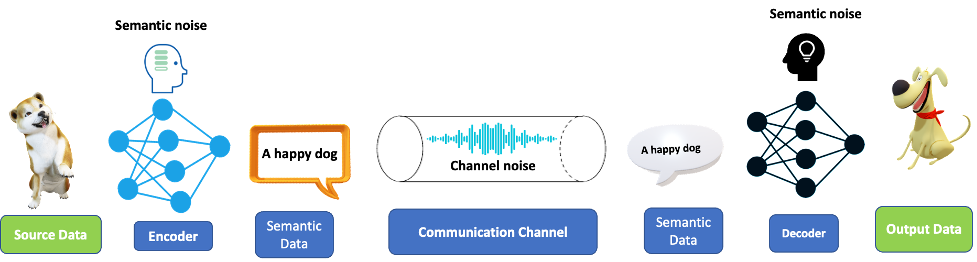}}
\caption{The basic blocks of a semantic communication}
\label{fig:SemComm_concept}
\end{figure*}

Semantic Communication (SemCom) is a novel communication model that focuses on transmitting only semantically-significant information through a communication channel (Fig.~\ref{fig:SemComm_concept})~\cite{getu2023making}. The objective of SemCom is to convey only the intended meaning behind a message as text or with model attributes, making communication more efficient by minimizing power usage, bandwidth consumption, and transmission delay. By reducing extraneous information that does not contribute to the message's meaning, SemCom optimizes information transmission and improves communication performance.  SemCom can maximize its energy-saving prospect by minimizing energy use in encoders and decoders. Our research presents a multi-objective loss metric function named energy-optimized semantic loss (EOSL) and thus offers a more robust and comprehensive SemCom system model. A holistic framework for deep learning based SemCom, including performance metrics and suitable AI architecture are crucial ~\cite{luo2022semantic}. In communication systems, semantic transformation loss refers to the degradation or alteration of the transmitted data's meaning or information content. This loss can arise from discrepancies in the interpretation of the data between the sender and receiver or because of errors occurring during transmission. The impact of semantic transformation loss may include misunderstandings, misinterpretations, or incomplete information, which could lead to inefficiencies or errors in the communication process.

Researchers attempt to address the issue of diverse information losses by employing a variety of metrics, including Structural Similarity Index Measure (SSIM), Word Error Rate (WER), Peak Signal-to-Noise ratio (PSNR), and Kullback–Leibler divergence (KID)~\cite{sun2013improved}. They focused on optimizing energy efficiency in SemCom using generative user selection and graph theory \cite{yang2022performance,lee2023energy}. However, they did not consider the correlation between energy consumption and semantic loss. Scholars explored federated learning as a multi-agent generative method for energy efficiency~\cite{kim2023green,zou2023wireless}. They employed diverse metrics like FLOP and training energy to gauge neural network energy consumption. In contrast, our approach directly measures CPU, GPU, and system utilization to accurately gauge energy consumption during inference, enhancing energy and efficiency comparisons. Our study reveals that informed model selection notably improves semantic efficiency while minimizing resource requirements. 

The unique contributions of this paper include:
\begin{itemize}
    \item Introduction of the novel \textbf{Energy Optimized Semantic Loss (EOSL)} guiding transformer model selection for substantial improvements in semantic efficiency without excessive computational and energy resources.
    
    \item Extensive benchmarking of resource utilization by transformer models, including CPU and GPU energy usage.
    
    \item Conducting various simulation studies to validate the efficiency of EOSL in choosing the optimal set of models for semantic encoding and decoding.
\end{itemize}

\section{Literature Review}

\begin{figure}[hbt!]
\centerline{\includegraphics[width=0.5\textwidth]{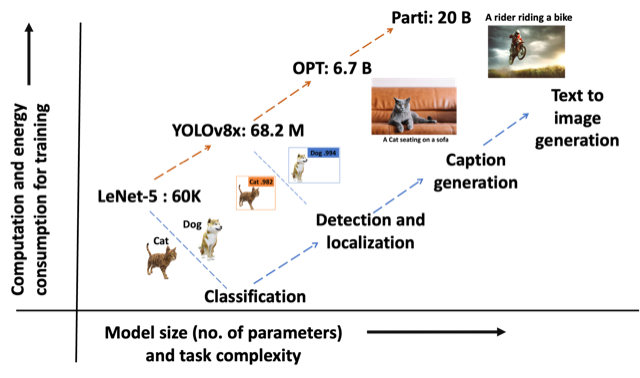}}
\caption{Evolution of complexity and training requirement of models}
\label{fig:SemComm_complexity}
\end{figure}

Deep learning models have grown significantly in size and computational requirements over time, driven by the need to perform more complex tasks that demand higher model complexity and larger data sets~\cite{hu2021model}. Early deep-learning models had only a few layers and limited parameters, primarily used for basic image and speech recognition. However, as the field has progressed, larger and more complex models have been developed to tackle more challenging problems, such as natural language processing, computer vision, and speech synthesis. Looking ahead to the near future, the trend of increasing model complexity and computational requirements are expected to continue. Fig.~\ref{fig:SemComm_complexity} shows LeNet~\cite{lecun1998gradient} using only 60k parameters for image classification, object detection using YOLOv8x~\cite{Jocher_YOLO_by_Ultralytics_2023} with 68 M parameters, OPT~\cite{zhang2022opt} for caption generation using 6.7 B, and Parti~\cite{yu2022scaling}, a text-to-image generation model by Google can scale up to 20 billion parameters. The well-known AlexNet architecture introduced in 2012 had around 60 million parameters, whereas modern state-of-the-art Large Language Models (LLM) like GPT-3 and EfficientNet can have billions of parameters. EfficientNet-B0, for instance, has only 5.3 million parameters but can achieve state-of-the-art accuracy on ImageNet with 6.4 times fewer FLOPs than the previous state-of-the-art model, while using 8.4 times less memory. While deep learning models have become faster to train with better hardware, their growing complexity still demands significant energy. This necessitates choosing efficient models that minimize environmental impact and enable deployment in resource-limited settings. MobileNet exemplifies energy-efficient design for mobile and similar contexts. The original MobileNet model had only 4.2 million parameters and could be trained with as little as 500,000 images, much smaller than other state-of-the-art models for image recognition. In an experiment,~\cite{garcia2019estimation} showed MobileNet as the most energy-efficient ConvNet choice under similar execution environments compared to Inception-V3 and DenseNet. EfficientNet is another energy-efficient deep-learning model that achieves state-of-the-art performance while using fewer parameters and less computation. It achieves this by using a novel compound scaling method that scales the model's depth, width, and resolution in a principled way. EfficientNet-B0 has only 5.3 million parameters and an energy efficiency of 4.6 billion operations per joule, which is much higher than other state-of-the-art models. Energy-efficient deep learning models like MobileNet and EfficientNet are ideal for resource-constrained environments as they achieve state-of-the-art performance with relatively low computational requirements and energy usage.

\section{Energy Optimized Semantic Loss}

Driven by the challenge of balancing semantic fidelity and energy efficiency in semantic communication, we introduce the ``Energy-Optimized Semantic Loss" (EOSL) function. This multi-objective metric captures both the semantic information loss and the energy requirements of the communication process, allowing for informed model selection and resource optimization. EOSL can be defined as follows.

\vspace{-0.6cm}
\begin{equation}
\small 
EOSL = \sum_{j=1}^{n} \left\{ \lambda_{sm}(N_{sm_j}) + \lambda_{lch}(L_{ch_j}) + \lambda_{e_{c}}(C_{e_j}) + \lambda_{e_{s}}(M_{e_j}) \right\}
\label{eq:EOSL}
\end{equation}

In this context, $n$ represents the total number of re-transmissions until the semantic noise requirement, $N_{sm_j} \leq N_{sm_{\text{thresh}}}$ is satisfied. The process of encoding, transmitting encoded messages, and decoding iterates until the condition is met. Here, $N_{sm_{\text{thresh}}}$ signifies the predetermined threshold for semantic noise. It is important to clarify that in this context, the EOSL is not employed as a training loss or regularization term. Its primary purpose is the selection of the most effective transformer model, serving either as an encoder or decoder, based on criteria that combine both semantic similarity and energy consumption. Consequently, it does not introduce any additional constraints or optimizations into the training procedure.

\begin{figure}[hbt!]
\centerline{\includegraphics[width=0.5\textwidth]{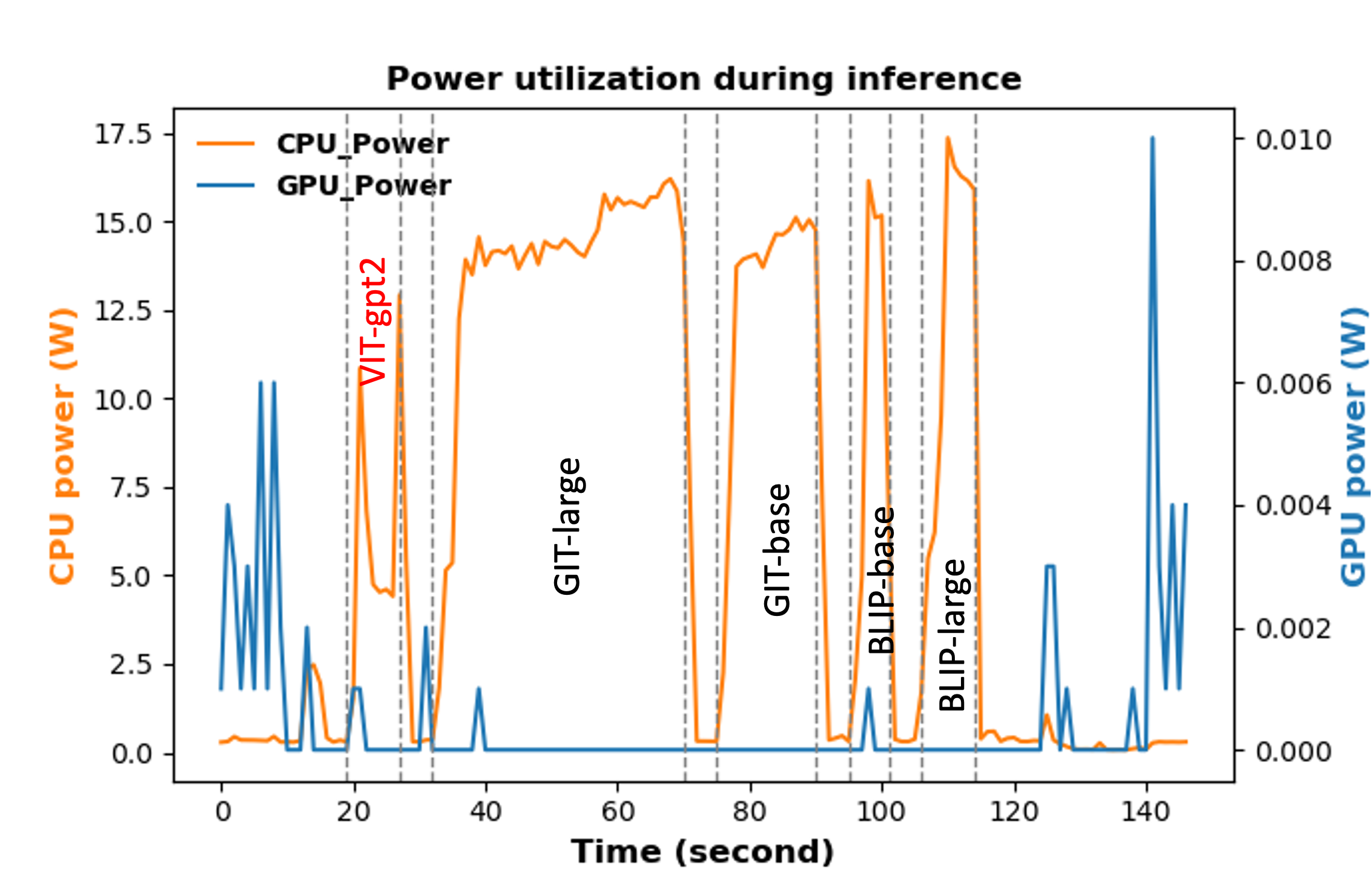}}
\caption{Resource utilization during inference }
\label{fig:tfm_comp}
\end{figure}

In (\ref{eq:EOSL}), EOSL is composed to combine four different types of factors that may impact efficient transmission. $N_{sm}$ represents a composite function of generic semantic or cognitive noise and its definition can vary depending on the type of semantic transformation applied, characterizing the disparity between the intended meaning of a message and its perceived meaning. $L_{ch}$ is defined as effective channel loss, analogous to traditional communication channels. The parameters $C_e$ and $M_e$ stand for ``Communication Energy" and ``Semantic Energy," respectively, and they are responsible for energy estimation. $N_{sm_{j}}$, $L_{ch_{j}}$, $M_{e_{j}}$, and $C_{e_{j}}$ represent the values of these variables during the $j$-th transmission attempt. We have also incorporated a set of four weight parameters (denoted as $\lambda$) to finely regulate the influence of each loss parameter within our model. Specifically, $\lambda_{sm}$ dictates the enforcement of semantic loss, while $\lambda_{lch}$ governs the effects of channel conditions, particularly the bit error probability encountered within the communication channel. Finally, $\lambda_{e_c}$ and $\lambda_{e_s}$ have been employed to calibrate the weighting of energy consumption for both transmitting and decoding/encoding the transmitted message. Let's denote the intended meaning of a message as $M_i$ and the perceived meaning as $M_p$. Then, the amount of semantic noise in the message can be quantified as:

\vspace{-0.4cm}
\begin{equation}
N_{sm} = f (M_i, M_p)    
\vspace{-0.2cm}
\end{equation}

Here, $f()$ is a function that measures the degree of similarity or dissimilarity between the intended and perceived meanings of the message. The larger the value of $f()$, the greater the amount of semantic noise in the message. One common way to compute $f()$ is by using a semantic similarity metric, which measures the degree of relatedness or likeness between two pieces of data (text, image, or media) based on their underlying meaning. The semantic similarity metric can be denoted as $S_{sm}(M_i, M_p)$. In current research, we experimented with two sets of similarity metrics: cosine similarity and structural similarity index measure (SSIM), to compare the original input message and the final output message. Cosine similarity ~\cite{wikipedia} between any two vectors is given by the  expression below:

\begin{equation}
\cos ({\bf A},{\bf B})= {{\bf A} {\bf B} \over \|{\bf A}\| \|{\bf B}\|} = \frac{ \sum_{i=1}^{n}{{\bf A}_i{\bf B}_i} }{ \sqrt{\sum_{i=1}^{n}{({\bf A}_i)^2}} \sqrt{\sum_{i=1}^{n}{({\bf B}_i)^2}} }
\end{equation}
where \textbf{A} and \textbf{B} are the two vectors being compared using cosine similarity. In addition, SSIM~\cite{1284395} can be used to measure  similarity between two images. It is based on the concept that the human eye is more sensitive to changes in structure than to changes in luminance or contrast. The mathematical formula for SSIM is defined as below:

\begin{equation}
  SSIM(x,y) = \frac{(2\mu_x\mu_y + C_1) + (2 \sigma _{xy} + C_2)} 
    {(\mu_x^2 + \mu_y^2+C_1) (\sigma_x^2 + \sigma_y^2+C_2)}
  \label{eq:SSMI}
\end{equation}
Parameters used in the above SSIM formula are:
\begin{itemize}
    \item $\mu_x$ the pixel sample mean of $x$
    \item $\mu_y$ the pixel sample mean of $y$
    \item $\sigma_x^2$ the variance of $x$
    \item $\sigma_y^2$ the variance of $y$
    \item $\sigma_{xy}$ covariance of $x$ and $y$
    \item $C_1$ and $C_2$ are two variables to stabilize the division with the weak denominator    
\end{itemize}

Here we can express the amount of semantic noise as:

\begin{equation}
N_{sm} = 1 - S_{sm} (M_i, M_p) \label{eq:Nsm}
\end{equation}
The specific form of the function $f()$ and the choice of semantic similarity metric may vary depending on the specific application and context in which semantic noise is being measured. Next, we consider the effects of channel noise on the accuracy of the received message. Given a probability of bit error $p_b$ caused by random channel noise and the average $E_b/N_0$ in a particular environment. Also, there is a probability $p_{f}$ of being in a deep fade, in which case all bits are lost (coincidentally the probability $0.5$ of bits in error). Given is the following average probability of bit error $\bar{p_b}$.

\begin{equation}
    \bar{p_b}=0.5p_{f}+p_b(1-p_{f})
\end{equation}
Then we create a channel loss metric to compare traditional and semantic communications by assuming a channel loss $L_{ch}$ analogous to the block error rate in the presence of a channel coding that can correct up to $t$ errors. 

\begin{equation}
L_{ch}=1-\sum_{i=0}^{t} {l\choose i} {\bar{p_b}}^i (1-{\bar{p_b}})^{l-i}
\end{equation}
where $l$ is the length of the packet in bits.
Next, we will compare the communication energy required for transmission using both traditional and semantic communication methods, highlighting the significant reduction in communication energy achieved with SemCom. The term $C_e$ represents the communication energy used in both traditional and semantic approaches. Finally, the semantic energy referred to as $M_e$, can be defined as the energy required for encoding or decoding semantic messages ($E_s$). EOSL normalizes both energies by dividing them by their maximum energy values $E_{c,max}$, and $E_{s,max}$ respectively among all available encoder/decoder options, See~(\ref{eq:EOSL_full}) below. We include the multipliers $\lambda$ to be able to tune the impact of the losses compared to the other terms. Hence the goal of our model will be to minimize EOSL which can be expressed as below:
\vspace{-0.2cm}
\begin{equation}
\begin{aligned}
EOSL &= \sum_{j=1}^{n} \left\{ \lambda_{sm}\Bigl(1 - S_{sm_j}(M_i, M_p)\Bigr) +  \lambda_{lch}\Bigl(L_{ch_j}\Bigr) + \right. \\
& \left. \lambda_{e_c}\Bigl(\frac{E_{c_j}}{E_{c,\text{max}}}\Bigr) + \lambda_{e_s}\Bigl(\frac{E_{s_j}}{E_{s,\text{max}}}\Bigr) \right\}
\end{aligned}
\label{eq:EOSL_full}
\end{equation}

EOSL will be used to compare SemCom encoders with each other and traditional communications, while the Decoder incurs even higher energy usage, discussed later.

\begin{figure}[hbt!]
\centerline{\includegraphics[width=0.5\textwidth]{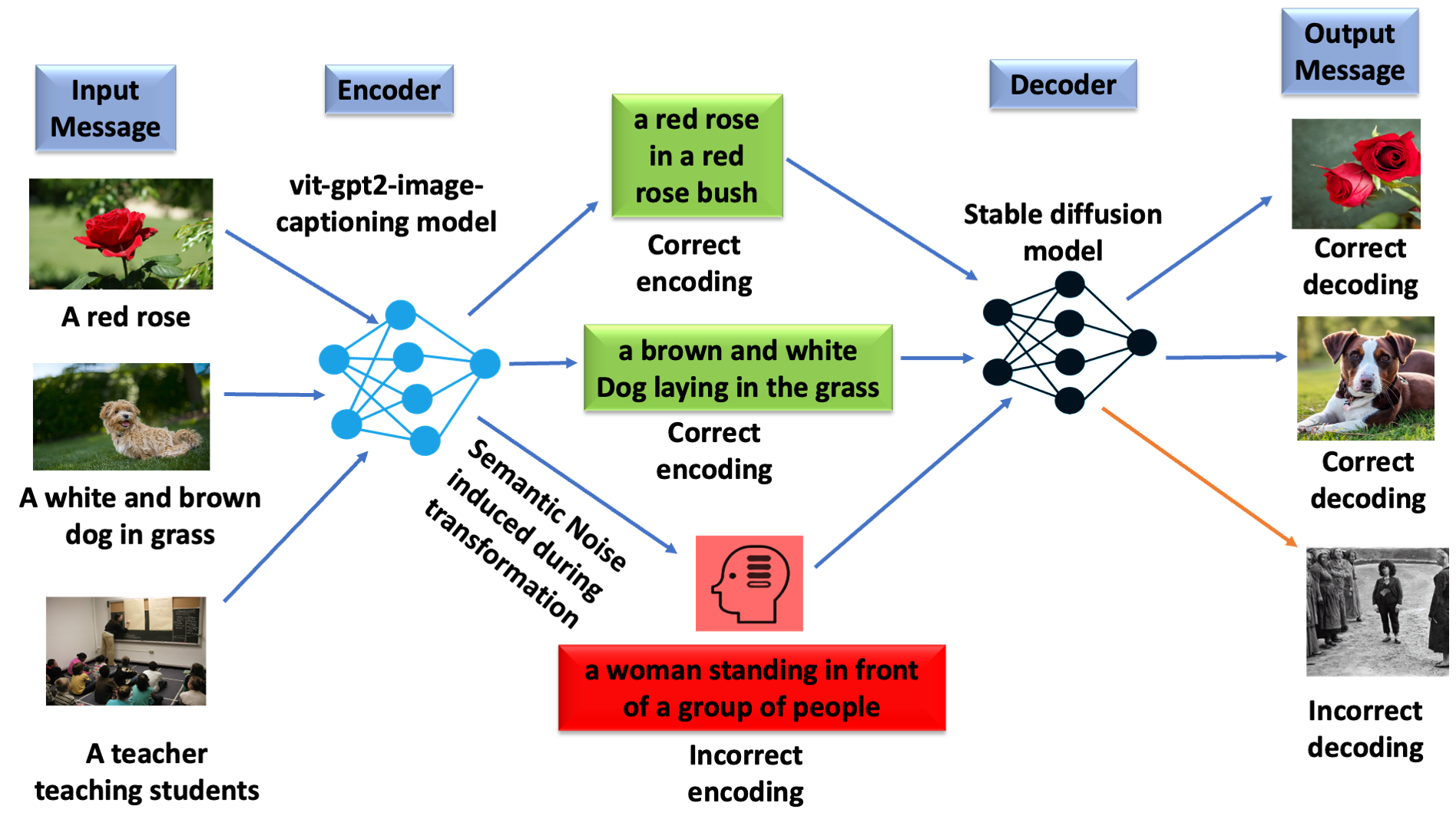}}
\caption{Encoder and decoder block design using image/text transformation and effect of semantic noise}
\label{fig:SemComm_EncDec}
\end{figure}

\section{Semantic Encoder and Decoder design}

We primarily designed the Encoder and Decoder system blocks and performed testing to exhibit and assess the semantic noise ($N_{sm}$).  We used pre-trained transformer-based model checkpoints~\cite{wolf2020transformers}  hosted in the Hugging Face public repository to design our semantic encoder. Transformers have emerged as a prominent neural network architecture, specifically designed to tackle sequence-to-sequence tasks encompassing machine translation, text summarizing, and question answering. Leveraging an attention mechanism, transformers excel in capturing intricate relationships among diverse segments within a sequence. This characteristic makes them applicable not only to text-based scenarios but also enables their utilization in cross-modal tasks such as text-to-image and image-to-text conversions. In text-to-image and image-to-text conversion tasks, Transformers exhibit their capability to grasp long-range dependencies. By establishing associations between textual descriptions and corresponding image pixels, transformers acquire the capability to generate visually coherent images aligned with the provided textual input. Consequently, the inherent ability of transformers to facilitate inter-modality conversion renders them a highly capable choice for constructing the encoder and decoder components of a semantic communication system. 

This transformer model transforms an image into text, to be transmitted via a communication channel. We evaluated several encoders. On the other end, we used another neural network model, the Stable Diffusion Model~\cite{rombach2022high} to design our semantic decoder for text to image. We followed a CUDA-enabled implementation, initially developed using CLIP (Contrastive Language-Image Pre-Training) by openAI~\cite{radford2021learning} and piped that to the CPU.

As illustrated in Fig.~\ref{fig:SemComm_EncDec}, we tested the semantic encoder and decoder with three images, and it successfully decoded the first two messages of “A red rose” and “A white and brown dog in grass”. After decoding, the semantics were preserved from the first two messages, so when they were again decoded using our semantic decoder, going from text to image, they were able to preserve the semantics of the input message. But the third message, which is a picture of “A teacher teaching students”, was incorrectly encoded by our semantic encoder; this is an example of semantic or cognitive noise, which occurred due to misinterpretation by the encoder. As a result, when this text was again decoded by our semantic decoder it was transformed into an image having different semantics.

\section{Experimental results}
Here we present two sets of results for (1) an encoder-based experiment and (2) an encoder and decoder-based experiment.
\subsection{Image-to-text encoding and EOSL-based model selection}
We have chosen five different caption generator transformer models to perform detailed energy benchmarking experiments during the image-to-text generation inference task. Specifically, we utilized the five encoder models `BLIP-base(Bootstrapping Language-Image Pre-training),' `GIT-base(Generative Image Transformer),' `GIT-large,' `BLIP-large,' and `VIT-GPT-2(Vision Transformer),' to convert the image into text. These models were run locally on an Apple M1 chipset MacBook Air with 8~GB memory and 256~GB storage using the MacOS Ventura operating system. It had a total of 8 cores (4 performance and 4 efficiency). We used a MacOS-based CLI utility `Powermetrics' to collect raw energy utilization data while performing individual model inferences.

\begin{figure*}[hbt!]
\centerline{\includegraphics[width=\textwidth]{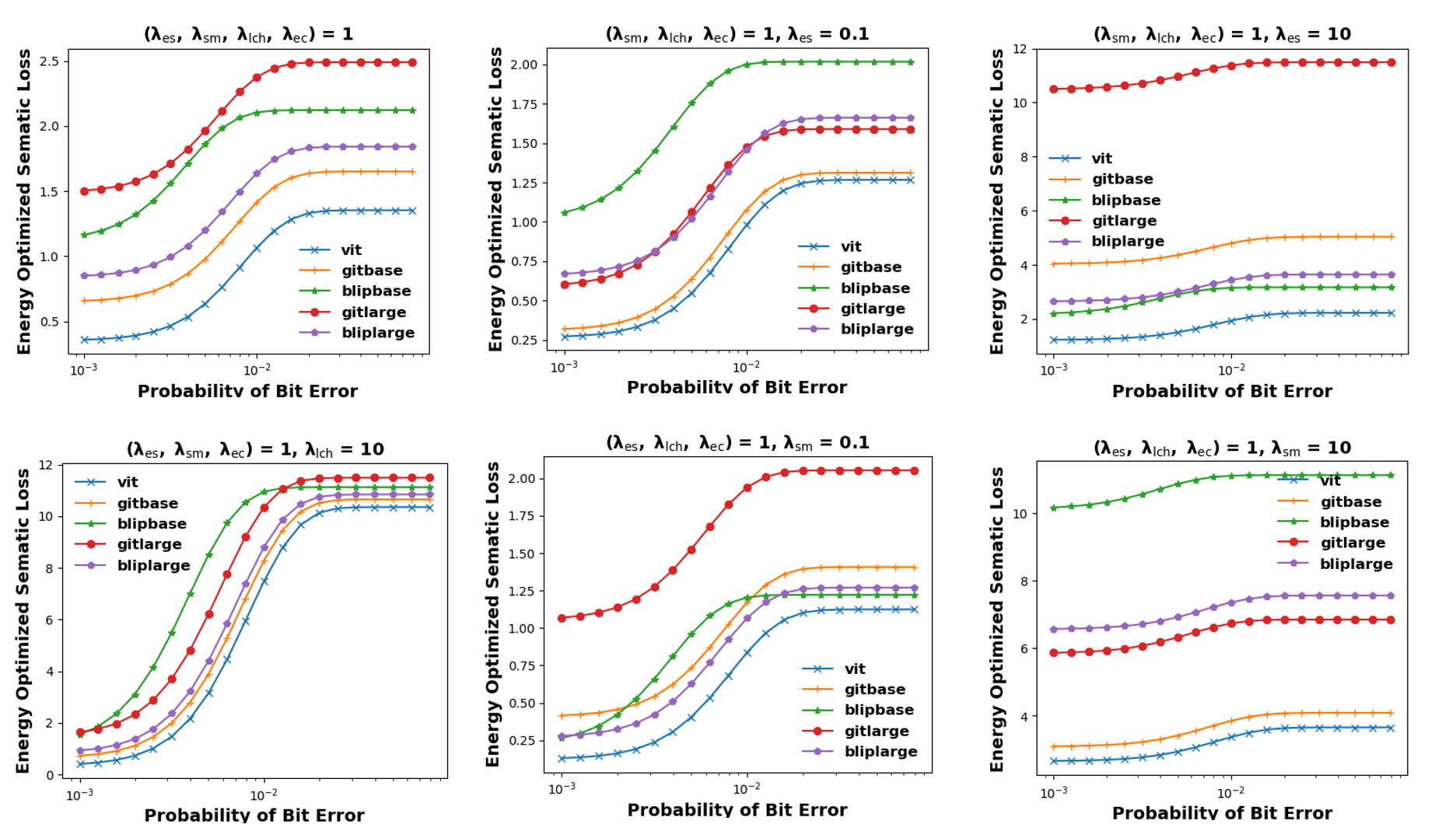}}
\caption{Changes of EOSL with the probability of bit error rate when using different values of $\lambda_{sm}$, $\lambda_{lch}$, $\lambda_{e_c}$, $\lambda_{e_s}$}
\label{fig:eosl_all}
\end{figure*}

\begin{table*}[htbp] 
\caption{Energy consumption during inference and EOSL values}
\centering
\footnotesize 
\setlength{\tabcolsep}{3pt} 
\begin{tabular}{|c|c|c|c|c|c|}
\hline
\textbf{Encoder} & \textbf{Total CPU Energy (J)} & \textbf{Total GPU Energy (J)} & \textbf{Total CPU Utilization (\%)} & \textbf{Semantic Noise} & \textbf{EOSL} \\
\hline
\textbf{VIT-GPT2} & \textbf{50.701} & \textbf{0.002} & \textbf{571.9} & \textbf{0.255} & \textbf{0.360} \\
\hline
BLIP-base & 60.922 & 0.001 & 513.4 & 1.000 & 1.164 \\
\hline
\textbf{GIT-base} & \textbf{197.442} & \textbf{0} & \textbf{1456.1} & \textbf{0.270} & \textbf{1.504} \\
\hline
BLIP-large & 105.095 & 0 & 746.3 & 0.635 & 0.659 \\
\hline
GIT-large & 524.718 & 0.001 & 3669.9 & 0.484 & 0.850 \\
\hline
\end{tabular}
\label{tab1}
\end{table*}

We selected a high-resolution (14 Mb) image of a dog as an input for the caption generation task (seen in Fig.~\ref{fig:dogs}) and used the 5 models to generate 5 different captions from the same image. We also defined a text description of the image `a brown dog running through grassy field', used as the semantics of this image. Those 5 generated captions were compared for text-based cosine similarity with our defined semantics. This gave us five different semantic similarity scores, from which we could calculate the semantic noise using equation~(\ref{eq:Nsm}). We also recorded CPU and GPU utilization performance data alongside the timestamps and duration for each model inference. All the consumption data was collected in 1-second intervals. Finally, we accumulated the most relevant parameters like total CPU and GPU energy (in Joules), CPU utilization \%, etc., then plotted them with respect to time in seconds on the x-axis as shown in Fig.~\ref{fig:tfm_comp}. The stop and start of each model inference for various transformers has been shown using grey-dotted vertical lines in the Fig.~\ref{fig:tfm_comp}.  As observed, larger models like GIT-large or BLIP-large had much higher energy footprints, but base models like VIT-GPT-2 or GIT-base consumed much less resources in terms of power and CPU utilization. The total energy consumed during an inference was obtained from the summation of instantaneous power as below:
\begin{equation}
E = \sum_{j=1}^{n} \sum_{i=1}^{m} P_{ij}\Delta t = \sum_{j=1}^{n} \sum_{i=1}^{m} P_{ij}
\end{equation}
Values are reported for $\Delta t=1$ sec, $P_{ij}$ is the instantaneous power at $i$-th second during $j$-th transmission. Also, when EOSL is plotted against bit error probability for various transformer models in Fig.~\ref{fig:eosl_all}, it can be observed that the VIT-GPT-2 model maintained the lowest EOSL with the increase in bit error probability. Moreover, given that these models primarily operate on CPUs, the utilization of GPU power is minimal in comparison to that of the CPU, as depicted in Fig.~\ref{fig:tfm_comp} and Table~\ref{tab1}. In this experiment, we assumed an average bit error probability of $0.001$, with a data rate of 143 Mbps (the rate per 20 MHz in IEEE 802.11ax), maximum admissible power of 1 Watt as regulated by FCC 15.247, and the average packet size of 1500 bytes as in a traditional communication system, while all the weight parameters are set to $\lambda$=1. Based on the results shown in Table~\ref{tab1}, VIT-GPT2, and GIT-base had lowest semantic noise; both are below $N_{\text{sm\_thresh}}=0.3$. However, we assume in the experiment $N_{sm_j} \leq N_{sm_{\text{thresh}}}$ is satisfied at $i=1$; no re-transmission was required.

\subsection{Image-to-text encoding and text-to-image decoding}
\begin{table*}[htbp] 
\caption{Encoder size and complexity with semantic efficiency}
\centering
\begin{tabular}{|c|c|c|c|c|c|c|c|c|}
\hline
\textbf{\textit{Encoder}} & \textbf{\textit{Decoder}} & \textbf{\textit{Size(Mb)}} & \textbf{\textit{Parameters(M)}} & \textbf{\textit{Cosine Similarity}} & \textbf{\textit{SSIM}} & \textbf{\textit{EOSL(using cosine)}} & \textbf{\textit{EOSL (using SSIM)}} \\
\hline
\textbf{GIT-base} & \multirow{5}{*}{Small-Stable-Diffusion-v0} & \textbf{673} & \textbf{177} & \textbf{0.878} & \textbf{0.654} & \textbf{0.321} & \textbf{0.123} \\
\cline{1-1} \cline{3-8}
{VIT-GPT2} & & 936 & 239 & 0.842 & 0.556 & 0.189 & 0.234 \\
\cline{1-1} \cline{3-8}
{BLIP-base} & & 943 & 247 & 0.837 & 0.530 & 0.267 & 0.321 \\
\cline{1-1} \cline{3-8}
{GIT-large} & & 1503 & 394 & 0.836 & 0.592 & 0.412 & 0.543 \\
\cline{1-1} \cline{3-8}
{BLIP-large} & & 1791 & 470 & 0.822 & 0.238 & 0.524 & 0.345 \\
\hline
\end{tabular}
\label{tab2}
\end{table*}

\begin{figure*}[hbt!]
\centerline{\includegraphics[width=\textwidth]{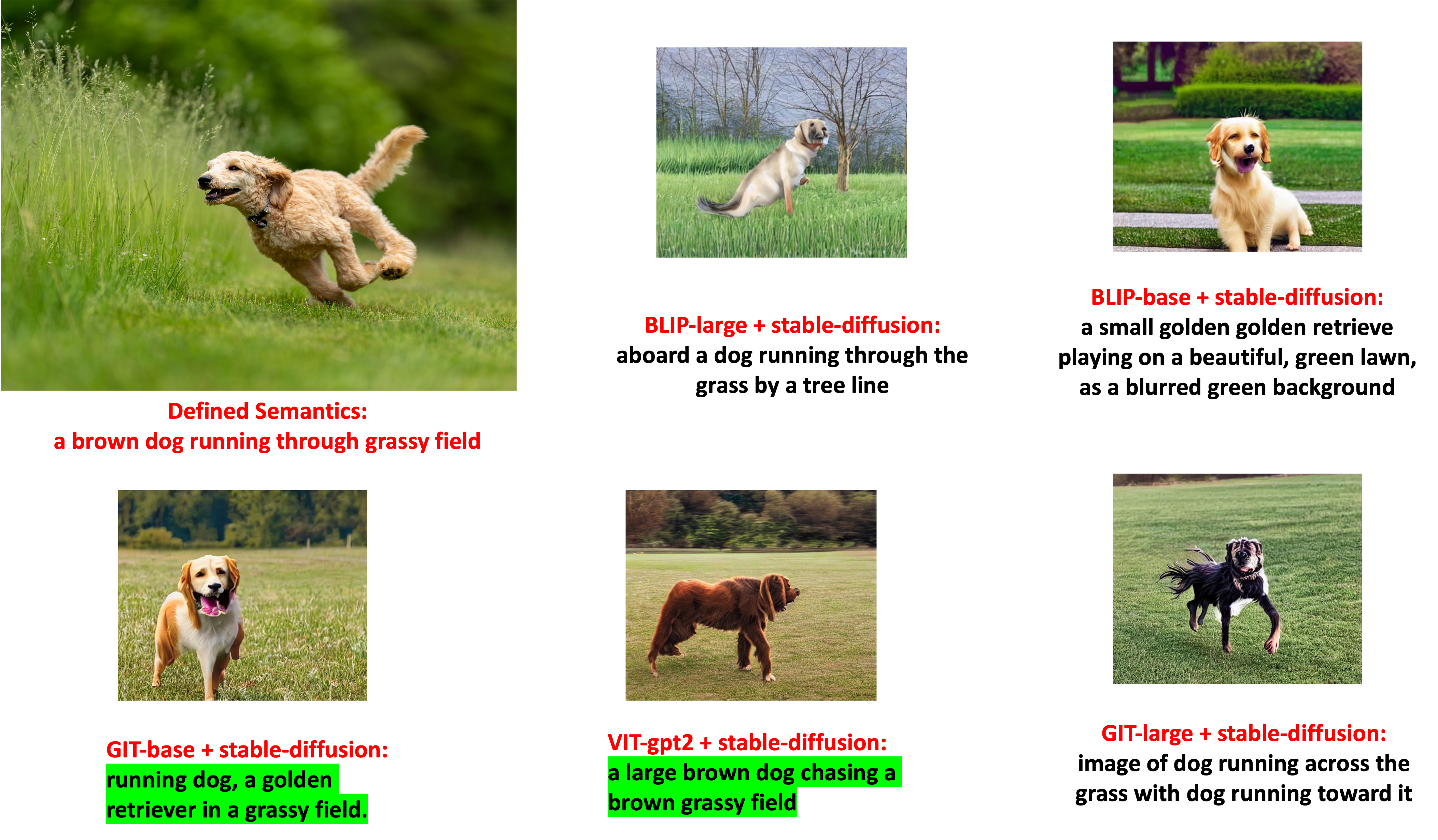}}
\caption{Illustration of semantic efficiency of different encoder models and images generated by diffusion}
\label{fig:dogs}
\end{figure*}

We conducted another experiment involving the transformation of a sample image to text and then from text to image using transformer models for both input and output stages. The same five encoder models, as mentioned in the first experiment, were used to convert the image into text. Additionally, in this case, a single decoder model, which is a text-to-image generator transformer named `Small-Stable-Diffusion-v0' was deployed for the reverse transformation, i.e. text to image generation. Fig.~\ref{fig:dogs} shows the main image, the semantics below it, and the text generated by all the five models are shown, along with the images generated from each text by stable diffusion model. Interestingly, our findings from Table~\ref{tab2} reveals that the similarity metrics are not dependent on or influenced by the sizes of the models utilized. This observation held when we repeated the experiment with images having their background removed, obtaining similar results. Remarkably, based on the outcomes of our current experiments, the `GIT-base' encoder model emerged as the most promising candidate, as it exhibited superior semantic efficiency across all types of similarity comparisons presented in Table~\ref{tab2}. This finding is notable considering that the `GIT-base' model is relatively smaller in size and possesses fewer hyper-parameters compared to larger and more complex alternatives such as `BLIP-Large' and `GIT-Large,' as shown in Table~\ref{tab2}. It should be noted that the semantic decoder model consumed approximately 40 times more energy than the 5 encoder models.  Text-to-image creation consumed an average of approximately 4~kJ which would heat 10 ml of water from room temperature to boiling.

\section{Conclusion and future work}
The experimental findings and $EOSL$-based selection algorithm presented herein demonstrate that selecting an optimized green semantic communication model yields notable enhancements in semantic efficiency without necessitating excessive computational and energy resources. The results show that an EOSL-based choice of encoder transformer models facilitates the semantic system to accomplish intricate tasks like semantic transformations (e.g., image-to-text conversions) with superior semantic efficiency while ensuring limited energy consumption, thus diverging from the trend shown in Fig.~\ref{fig:SemComm_complexity} that increasingly complex tasks tend to require more complex models. In semantic communication systems, energy spent on message transmission can decrease, and energy used for transformations increase with larger transformer models. Our experiments confirm that selecting energy-efficient, yet semantically intelligent encoder and decoder models can effectively lessen the energy demands of these transformations.

This work can be extended toward energy benchmarking that encompasses a broader range of platforms, from GPU clusters to Raspberry PIs. Text-to-image decoders can be developed that do not have 40 times the energy of encoders. Research would cover a complete dataset rather than just a single sample and involve tuning the parameters and structures of particular transformers themselves for better $EOSL$ and energy trade-offs. Additionally, enhancing resource parameters with more granularity is feasible. We intend to create an energy-efficient semantic inference simulator that will comprehensively compare semantic capabilities and energy efficiency among various encoders and decoders. It will incorporate standard or customizable similarity functions and detailed energy consumption benchmarks. This approach would identify the most energy-optimized candidates, aligned to achieve environmentally friendly end-to-end semantic communication.

\bibliography{reference}

\begin{thebibliography}{10}
\providecommand{\url}[1]{#1}
\csname url@samestyle\endcsname
\providecommand{\newblock}{\relax}
\providecommand{\bibinfo}[2]{#2}
\providecommand{\BIBentrySTDinterwordspacing}{\spaceskip=0pt\relax}
\providecommand{\BIBentryALTinterwordstretchfactor}{4}
\providecommand{\BIBentryALTinterwordspacing}{\spaceskip=\fontdimen2\font plus
\BIBentryALTinterwordstretchfactor\fontdimen3\font minus \fontdimen4\font\relax}
\providecommand{\BIBforeignlanguage}[2]{{%
\expandafter\ifx\csname l@#1\endcsname\relax
\typeout{** WARNING: IEEEtran.bst: No hyphenation pattern has been}%
\typeout{** loaded for the language `#1'. Using the pattern for}%
\typeout{** the default language instead.}%
\else
\language=\csname l@#1\endcsname
\fi
#2}}
\providecommand{\BIBdecl}{\relax}
\BIBdecl

\bibitem{getu2023making}
T.~M. Getu, G.~Kaddoum, and M.~Bennis, ``Making sense of meaning: A survey on metrics for semantic and goal-oriented communication,'' \emph{IEEE Access}, 2023.

\bibitem{luo2022semantic}
X.~Luo, H.-H. Chen, and Q.~Guo, ``Semantic communications: Overview, open issues, and future research directions,'' \emph{IEEE Wireless Communications}, vol.~29, no.~1, pp. 210--219, 2022.

\bibitem{sun2013improved}
K.~Sun, Y.~Ji, L.~Rui, and X.~Qiu, ``An improved method for measuring concept semantic similarity combining multiple metrics,'' in \emph{2013 5th IEEE International Conference on Broadband Network \& Multimedia Technology}.\hskip 1em plus 0.5em minus 0.4em\relax IEEE, 2013, pp. 268--272.

\bibitem{yang2022performance}
Z.~Yang, M.~Chen, Z.~Zhang, C.~Huang, and Q.~Yang, ``Performance optimization of energy efficient semantic communications over wireless networks,'' in \emph{2022 IEEE 96th Vehicular Technology Conference (VTC2022-Fall)}.\hskip 1em plus 0.5em minus 0.4em\relax IEEE, 2022, pp. 1--5.

\bibitem{lee2023energy}
H.~Lee, J.~Park, S.~Kim, and J.~Choi, ``Energy-efficient downlink semantic generative communication with text-to-image generators,'' \emph{arXiv preprint arXiv:2306.05041}, 2023.

\bibitem{kim2023green}
M.~Kim, W.~Saad, M.~Mozaffari, and M.~Debbah, ``Green, quantized federated learning over wireless networks: An energy-efficient design,'' \emph{IEEE Transactions on Wireless Communications}, 2023.

\bibitem{zou2023wireless}
H.~Zou, Q.~Zhao, L.~Bariah, M.~Bennis, and M.~Debbah, ``Wireless multi-agent generative ai: From connected intelligence to collective intelligence,'' \emph{arXiv preprint arXiv:2307.02757}, 2023.

\bibitem{hu2021model}
X.~Hu, L.~Chu, J.~Pei, W.~Liu, and J.~Bian, ``Model complexity of deep learning: A survey,'' \emph{Knowledge and Information Systems}, vol.~63, pp. 2585--2619, 2021.

\bibitem{lecun1998gradient}
Y.~LeCun, L.~Bottou, Y.~Bengio, and P.~Haffner, ``Gradient-based learning applied to document recognition,'' \emph{Proceedings of the IEEE}, vol.~86, no.~11, pp. 2278--2324, 1998.

\bibitem{Jocher_YOLO_by_Ultralytics_2023}
\BIBentryALTinterwordspacing
G.~Jocher, A.~Chaurasia, and J.~Qiu, ``{YOLO by Ultralytics},'' Jan. 2023. [Online]. Available: \url{https://github.com/ultralytics/ultralytics}
\BIBentrySTDinterwordspacing

\bibitem{zhang2022opt}
S.~Zhang, S.~Roller, N.~Goyal, M.~Artetxe, M.~Chen, S.~Chen, C.~Dewan, M.~Diab, X.~Li, X.~V. Lin, T.~Mihaylov, M.~Ott, S.~Shleifer, K.~Shuster, D.~Simig, P.~S. Koura, A.~Sridhar, T.~Wang, and L.~Zettlemoyer, ``Opt: Open pre-trained transformer language models,'' 2022.

\bibitem{yu2022scaling}
J.~Yu, Y.~Xu, J.~Y. Koh, T.~Luong, G.~Baid, Z.~Wang, V.~Vasudevan, A.~Ku, Y.~Yang, B.~K. Ayan, B.~Hutchinson, W.~Han, Z.~Parekh, X.~Li, H.~Zhang, J.~Baldridge, and Y.~Wu, ``Scaling autoregressive models for content-rich text-to-image generation,'' 2022.

\bibitem{garcia2019estimation}
E.~Garc{\'\i}a-Mart{\'\i}n, C.~F. Rodrigues, G.~Riley, and H.~Grahn, ``Estimation of energy consumption in machine learning,'' \emph{Journal of Parallel and Distributed Computing}, vol. 134, pp. 75--88, 2019.

\bibitem{wikipedia}
\BIBentryALTinterwordspacing
``{Cosine similarity},'' 2023. [Online]. Available: \url{https://en.wikipedia.org/wiki/Cosine\_similarity}
\BIBentrySTDinterwordspacing

\bibitem{1284395}
Z.~Wang, A.~Bovik, H.~Sheikh, and E.~Simoncelli, ``Image quality assessment: from error visibility to structural similarity,'' \emph{IEEE Transactions on Image Processing}, vol.~13, no.~4, pp. 600--612, 2004.

\bibitem{wolf2020transformers}
T.~Wolf, L.~Debut, V.~Sanh, J.~Chaumond, C.~Delangue, A.~Moi, P.~Cistac, T.~Rault, R.~Louf, M.~Funtowicz \emph{et~al.}, ``Transformers: State-of-the-art natural language processing,'' in \emph{Proceedings of the 2020 conference on empirical methods in natural language processing: system demonstrations}, 2020, pp. 38--45.

\bibitem{rombach2022high}
R.~Rombach, A.~Blattmann, D.~Lorenz, P.~Esser, and B.~Ommer, ``High-resolution image synthesis with latent diffusion models,'' in \emph{Proceedings of the IEEE/CVF Conference on Computer Vision and Pattern Recognition}, 2022, pp. 10\,684--10\,695.

\bibitem{radford2021learning}
A.~Radford, J.~W. Kim, C.~Hallacy, A.~Ramesh, G.~Goh, S.~Agarwal, G.~Sastry, A.~Askell, P.~Mishkin, J.~Clark \emph{et~al.}, ``Learning transferable visual models from natural language supervision,'' in \emph{International conference on machine learning}.\hskip 1em plus 0.5em minus 0.4em\relax PMLR, 2021, pp. 8748--8763.

\end{thebibliography}
\bibliographystyle{IEEEtran}

\end{document}